\pdfoutput=1

\documentclass[11pt]{article}

\usepackage{acl}

\usepackage{times}
\usepackage{latexsym}

\usepackage[T1]{fontenc}

\usepackage[utf8]{inputenc}

\usepackage{microtype}

\usepackage{graphicx}
\usepackage{epstopdf}
\usepackage{hyperref}
\usepackage{xstring}
\usepackage{color}
\usepackage{amsmath}
\usepackage{amssymb}
\usepackage{multirow}
\usepackage{makecell}
\usepackage{subcaption}

\def\tabref#1{Table~\ref{#1}}
\def\figref#1{Figure~\ref{#1}}
\def\equref#1{Equation~(\ref{#1})}

\def\appref#1{Appendix~\ref{#1}}

\title{Visual Recipe Flow: A Dataset for Learning Visual State Changes \\ of Objects with Recipe Flows}

\author{
      Keisuke Shirai$^1$ \quad
      Atsushi Hashimoto$^2$ \quad
      Taichi Nishimura$^1$ \quad
      Hirotaka Kameko$^1$ \quad \\
      \textbf{Shuhei Kurita}$^3$ \quad
      \textbf{Yoshitaka Ushiku}$^2$ \quad
      \textbf{Shinsuke Mori}$^1$ \\
      $^1$Kyoto University \quad
      $^2$OMRON SINIC X Corporation \quad
      $^3$RIKEN AIP, JST PRESTO \\
      \{shirai.keisuke.64x,nishimura.taichi.43x\}@st.kyoto-u.ac.jp \\ \quad
      \{atsushi.hashimoto,yoshitaka.ushiku\}@sinicx.com \\ \quad
      \{kameko,forest\}@i.kyoto-u.ac.jp \quad
      shuhei.kurita@riken.jp
}
      
\begin{document}
\maketitle

\begin{abstract}
We present a new multimodal dataset called Visual Recipe Flow, which enables us to learn each cooking action result in a recipe text. The dataset consists of object state changes and the workflow of the recipe text. The state change is represented as an image pair, while the workflow is represented as a recipe flow graph (r-FG). The image pairs are grounded in the r-FG, which provides the cross-modal relation. With our dataset, one can try a range of applications, from multimodal commonsense reasoning and procedural text generation.
\end{abstract}

\section{Introduction}
\begin{figure}[t]
    \centering
    
    \includegraphics[scale=0.345]{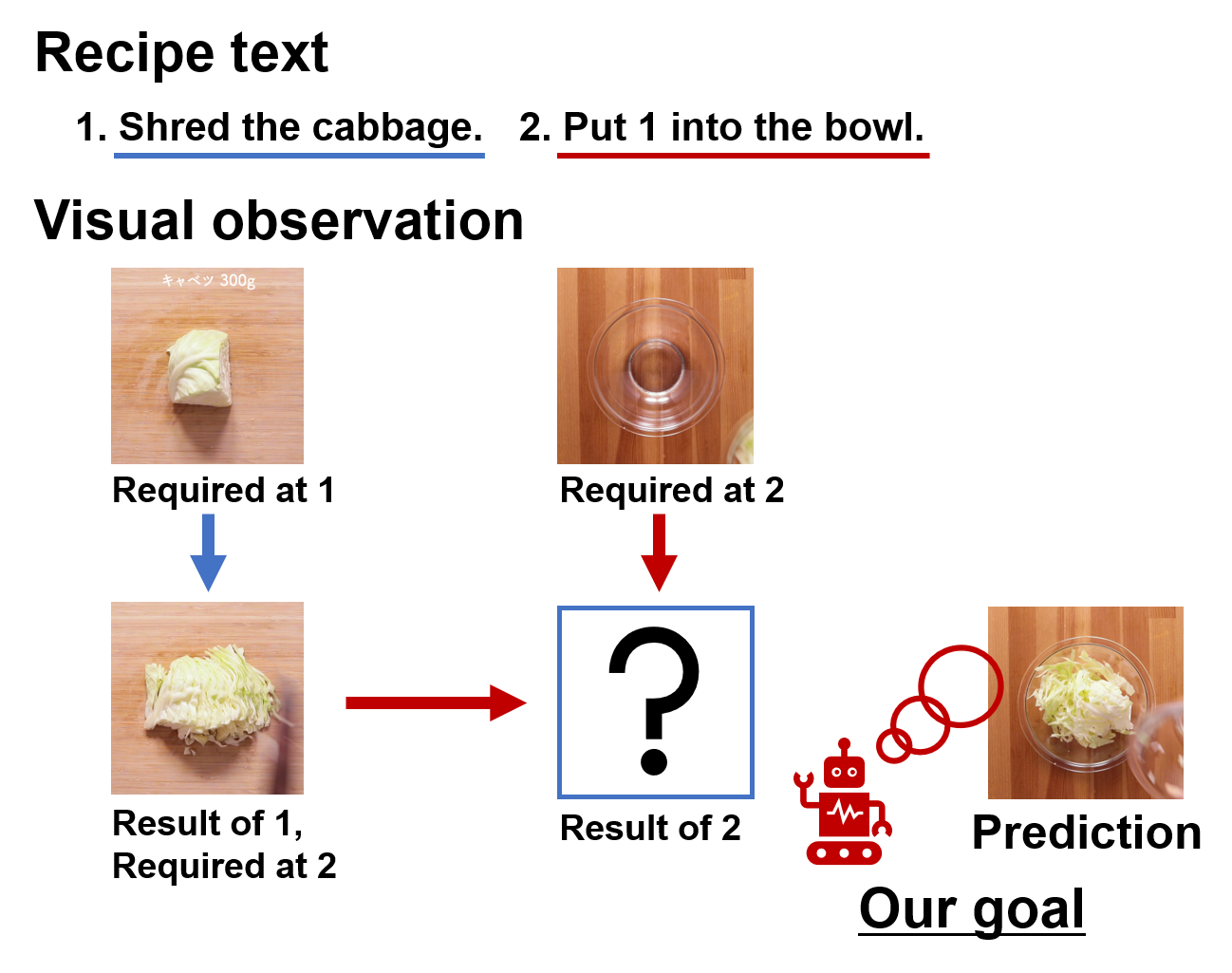}
    \caption{Our goal is to build an agent that tracks object state changes and predicts what observations can be obtained by cooking actions.}
    \label{fig:introduction-example}
\end{figure}

Our aim is to track how foods are processed and changed toward the final food product by each cooking action given a recipe text. This requires some knowledge of the actions: what foods and actions are involved and how the action changes them. Skilled chefs can easily imagine these action effects while understanding the required foods. We are interested in building an autonomous agent endowed with this ability, as illustrated in \figref{fig:introduction-example}. This example involves two cooking actions, and the agent imagines the second action result: the shredded cabbage in the bowl. This also implicates the food requirement: the shredded cabbage produced by the previous action. The prediction for the required foods and action results is indeed a natural ability for humans when they cook something. Thus, this is also crucial for intelligent autonomous agents to understand recipe texts.

Previous work on this line of research provided visual annotation for each cooking instruction~\citep{nishimura2020visual,pan2020multimodal}. \citet{nishimura2020visual} attached an image with bounding boxes of objects to each instruction, while \citet{pan2020multimodal} split an instruction into sentences and attached frames to each sentence. However, their annotations are often insufficient to predict the action result for each object. A typical case is an instruction in a sentence that directs multiple actions. For example, the instruction of ``slice the tomato and put it into the bowl'' produces two action results: the sliced tomato and that put in the bowl. Therefore, an instruction-wise visual annotation is insufficient for our task, and action-wise visual annotation is required. Preparing a more dense visual annotation is one straightforward way to handle this case.

\begin{figure*}[t]
    \centering
    \includegraphics[scale=0.251]{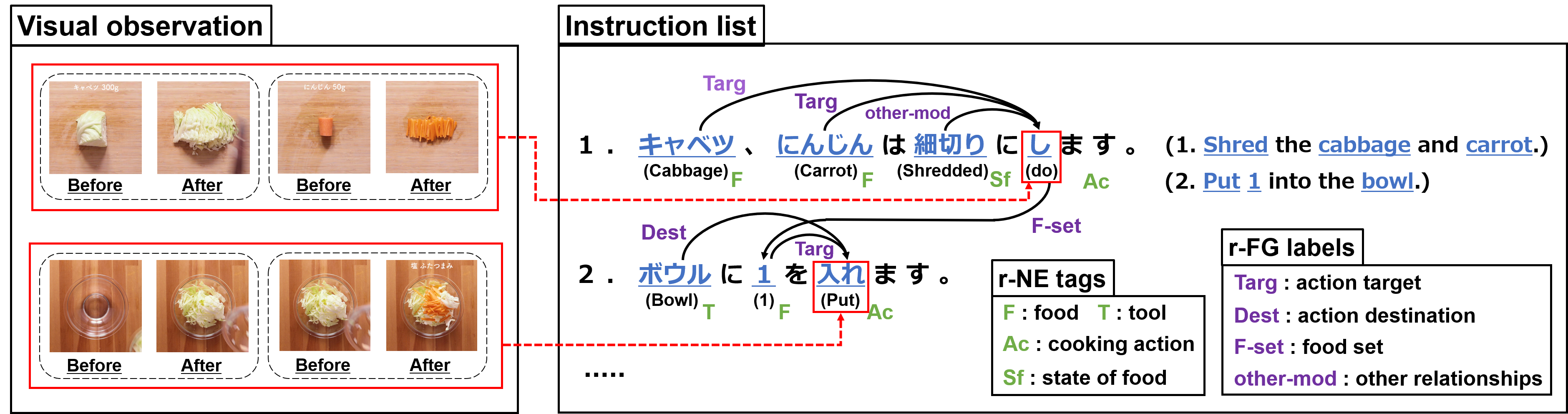}
    \caption{Example of our dataset. A pair of images in the visual observation corresponds to the states of object before and after a cooking action. They are grounded in the action in the instruction list. The black solid arrows denote recipe flows, which describe the relationships between expressions (e.g., cooking actions, foods, and tools).}
    \label{fig:dataset-example}
\end{figure*}

Toward the realization of an agent that predicts the result of each action, we introduce a new multimodal dataset called Visual Recipe Flow (VRF). The dataset consists of object\footnote{In our work, object refers to food or tool.} state changes caused by every action and the workflow of the text. The change is given as an image pair, while the workflow is given in the format of recipe flow graph (r-FG)~\citep{mori2014flow}. Each image pair is grounded in the r-FG, which gives the cross-modal relation. \figref{fig:dataset-example} shows an example of our dataset. We focus on recipe text involving various cooking actions, foods, and state changes, which is one of the representatives of procedural texts.

Understanding these texts by tracking object state changes is one of the recent trends~\citep{dalvi2018tracking,bosselut2018simulating,tandon2020dataset,nishimura2021state,papadopoulos2022learning}. Our work also contributes to this line of research. Since images directly express object appearances in the real world~\citep{isola2015discovering,zhang2021mirecipe}, our dataset would provide rich information for the changes.
The sequential nature of our dataset can also be used to test the reading comprehension ability of large-scale language models~\citep{srivastava2022beyond}. Furthermore, since our dataset has arbitrary interleaved visual and textual annotations, it is also possible to evaluate the few-shot capability of vision-language models on such data~\citep{alayrac2022flamingo}.

\section{The VRF dataset}\label{sec:dataset}
The Visual Recipe Flow (VRF) dataset is a new multimodal dataset. It provides visual annotations for objects in a recipe text before and after a cooking action. We identify expressions including the action in the text by using recipe named entities (r-NEs)~\citep{mori2014flow}, which can be extended to other procedural tasks. Based on the r-NEs, it also provides a representation of the recipe workflow as a recipe flow graph (r-FG)~\citep{mori2014flow}. In this section, we first explain the overview of the r-FG and then introduce our visual annotation.

\subsection{Recipe flow graph (r-FG)}
The r-FG represents the cooking workflow of a recipe text. It consists of a set of recipe flows. The recipe flow is expressed as a directed edge that takes two r-NEs as the starting and ending vertices. It also has a label that describes the relationship between them. It connects one cooking action with the next and expresses its dependencies. For example, in \figref{fig:dataset-example}, the first action is connected with the second one, which means that the second action requires the products of the first action: shredded cabbage and carrot. This helps us to identify what foods are required for the actions. The annotation has the flows from the ingredient lists to track foods from raw ingredients~\citep{nishimura2021state}, which allows us to convert the r-FG into cooking programs~\citep{papadopoulos2022learning}.

\subsection{Visual annotation}
Our visual annotation is given as an extension of the r-FG. Each annotation consists of a pair of images which represent object state change by the action. Each image pair is linked with the action in the r-FG. In some cases, a single action can require multiple objects and change their states. Our annotation provides an image pair to all of these state changes. In \figref{fig:dataset-example}, for example, the first action is linked with two image pairs because it induces the state changes of two objects: cabbage and carrot. This dense annotation would help develop autonomous cooking agents because these images provide visual clues for each action.

\section{Annotation standards}\label{sec:annotation-standards}
\begin{figure}[t]
    \centering
    \includegraphics[scale=0.32]{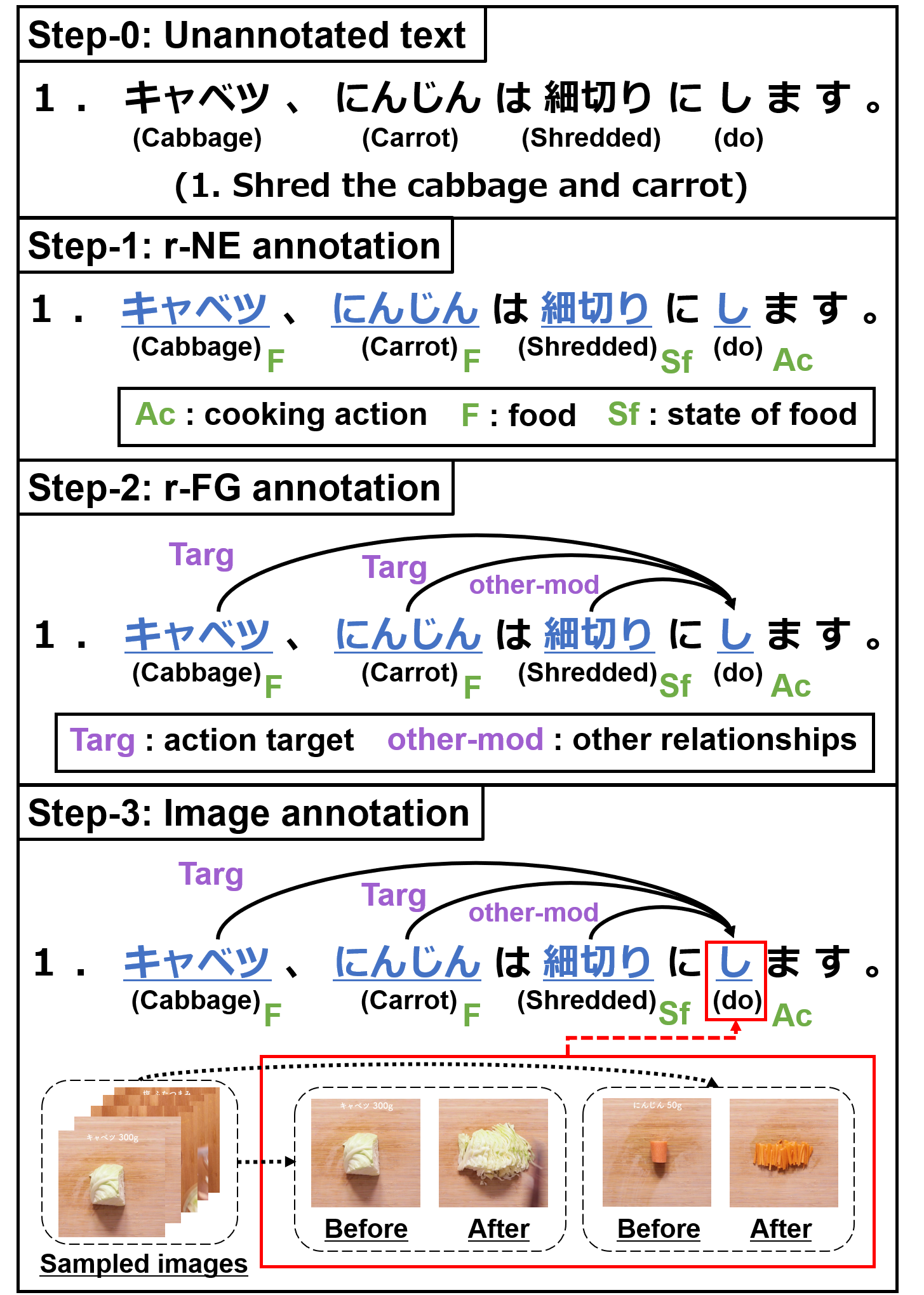}
    \caption{Example of annotation process for a single instruction. The instruction is sequentially annotated with r-NE tags, recipe flows, and images.}
    \label{fig:annotation-steps}
\end{figure}

In this section, we describe our annotation standards. The annotation consists of three steps in order: (i) r-NE annotation, (ii) r-FG annotation, and (iii) image annotation. Each recipe has an ingredient list, an instruction list, and a cooking video. \figref{fig:annotation-steps} shows an example of the annotations.

\paragraph{r-NE annotation.} 
First, we annotated words in the ingredient and instruction lists with r-NE tags\footnote{We segmented sentences into words beforehand by using a Japanese tokenizer, KyTea~\citep{neubig2011pointwise}, because words in a Japanese sentence are not typically separated by whitespace.}. We used the eight types of r-NE tags, following \citet{mori2014flow}.  See \appref{sec:r-NE-r-FG-stats} for details. 

\paragraph{r-FG annotation.}
Second, we annotated the r-NEs in the first step with the r-FG. We used the 13 types of r-FG labels, following \citet{maeta2015framework}. 
See \appref{sec:r-NE-r-FG-stats} for details. 

\paragraph{Image annotation.}
Third, we annotated object states with images, sampled at $3$ frames per second from the videos. Each object required for any cooking action is annotated as a pair of frames of states before and after the action. When there are multiple suitable frames, we prioritize the one based on the visual clarity of the object. In some cases, objects are always heavily covered by human hands or abbreviated from the video. We treat them as missing data.

\section{Annotation results}\label{sec:annotation-results}
This section first describes our annotation process and the statistics for the annotation results. It then investigates the dataset quality and finally assesses our dataset by conducting experiments.

\subsection{Annotation process}
We started by collecting recipes and cooking videos since the existing r-FG datasets~\citep{mori2014flow,yamakata2020english} are not necessarily associated with the videos. We collected $200$ recipes in Japanese and videos from the Kurashiru website\footnote{\url{https://www.kurashiru.com}, accessed on 2021/12/14.}. In the video, each cooking process is recorded in detail by a fixed camera. Thus, we can annotate the object states with a fixed viewpoint. Considering the future cooking agent developments, we focused on salad recipes, in which the procedures are simple but still contain $89$ unique expressions for cooking action and $275$ unique ingredients. 

We asked one Japanese annotator, familiar with the r-NE and r-FG, to annotate the recipes. However, filling spreadsheets manually~\citep{mori2014flow} is heavy, and it also might cause unexpected annotation errors. Therefore, we developed a web interface to help the annotation. The interface supports all three annotation steps. With this interface, the annotator can annotate recipes with r-NE tags, r-FG labels, and images by simple mouse operations. An illustration of the interface is provided in \appref{sec:web-interface}. The whole annotation took $120$ hours. 

In the annotation collection process, we created annotation guidelines to check annotation errors and reproduce high-quality annotations by another annotator. Starting with a draft, we iteratively revised the guidelines when the first 10, 20, and 50 recipe annotations were finished. In the verification process, we shared the guidelines and three annotation examples with the second annotator.

\subsection{Statistics}
The recipes contained $1,701$ ingredients, $1,077$ instructions, and $33,400$ words in total. The average number of ingredients and instructions per recipe was $8.51$ and $5.31$, respectively. The r-NE annotation resulted in $11,686$ r-NEs, while the r-FG annotation resulted in $11,291$ recipe flows. We provide the detailed statistics for them in \appref{sec:r-NE-r-FG-stats}. 

\tabref{tab:image-annotation} shows the statistics for the image annotation results. We annotated $3,705$ objects in the r-FGs with images. Among them, $2,551$ had both pre-action and post-action images, $485$ had only a post-action image, $72$ had only a pre-action image, and $597$ had no image. In total, $5,659$ images ($3,824$ unique images) were used. 

\subsection{Dataset quality}
To investigate the correctness and consistency of the annotation results, we asked another annotator to re-annotate $10$ recipes, which were randomly sampled from the collected recipes and contained $623$ named entity tags, $616$ recipe flows, and $199$ visual state changes. We then measured the inter-annotator agreements in precision, recall, and F-measure. The agreements were calculated between the two sets of annotations by taking the first one as the ground truth. 

\tabref{tab:annotator-agreement} lists the results. The F-measure for the r-NE was $98.40$, which was almost perfect agreement. The F-measure for the r-FG was $86.11$, which was also quite high considering that all the r-NEs were presented as candidate vertices. The F-measure for the images was $72.80$, which was smaller than the former steps. However, this was still high, considering that annotation differences in the former steps affected this step. 

\begin{table}[t]
    \centering
    \begin{tabular}{c|c|r}\hline
        \multicolumn{2}{c|}{Annotated image} & \multirow{3}{*}{\makecell{\# objects}} \\\cline{1-2}
        \multirow{2}{*}{\makecell{Pre-action \\ state}} & \multirow{2}{*}{\makecell{Post-action \\ state}} & \\
        & & \\\hline
         & & 597 \\
        \checkmark & & 72 \\
         & \checkmark & 485 \\
        \checkmark & \checkmark & 2,551 \\\hline
        \multicolumn{2}{c|}{Total} & 3,705 \\\hline
    \end{tabular}
    \caption{Statistics for the image annotation results. Objects have image annotation of a pre-action or post-action state if it is checked.}

    \label{tab:image-annotation}
\end{table}
\begin{table}[t]
    \centering
    \begin{tabular}{l|c|c|c}\hline
        \multicolumn{1}{c|}{Annotation} & Precision & Recall & F-measure \\\hline
        r-NE & 97.93 & 98.88 & 98.40 \\
        r-FG & 86.18 & 86.04 & 86.11 \\
        Image & 75.13 & 70.60 & 72.80 \\\hline
    \end{tabular}
    
    \caption{Inter-annotator agreements of the annotations.}
    \label{tab:annotator-agreement}
\end{table}
\subsection{Experiments}\label{sec:experiments}
We conducted multimodal information retrieval experiments to assess our dataset. The experiments aimed to find a correct post-action image from a set of candidate images by using the cooking action verb and pre-action image information. We used a joint embedding model~\citep{miech2019howto100m} and briefly explain the calculation here\footnote{See details in \appref{sec:appendix-model}}. We calculated a vector for an estimated post-action object state from the action verb and pre-action image information. This vector is mapped into a shared embedding space. On the other hand, the candidate post-action images are mapped into vectors and mapped them into the embedding space. We searched for the correct post-action image from the estimated post-action state based on their similarities in the embedding space. 

Our model was trained with different input configurations. We used the Recall@5 (R@5) and the median rank (MedR) as evaluation metrics. \tabref{tab:experimental-results} shows the results. The second and third lines' scores show that the image provides more information than the text. The fourth line's scores imply that the textual and visual modalities provide different information, and using them together is more effective. These results demonstrate that the visual modality provides critical information for finding post-action images. These also indicate the usefulness of our annotation.

\begin{table}[t]
    \centering
    \begin{tabular}{c|c|c|c} \hline
        \multicolumn{2}{c|}{Used input} & \multirow{3}{*}{R@5 ($\uparrow$)} & \multirow{3}{*}{MedR ($\downarrow$)} \\\cline{1-2}
        \multirow{2}{*}{\makecell{Action \\ verb}} & \multirow{2}{*}{\makecell{Pre-action \\ image}} & & \\
        & & & \\\hline
         & & ~~2.37 & 149.00 \\
        \checkmark & & 21.24 & ~~26.70 \\
         & \checkmark & 33.77 & ~~12.60 \\
        \checkmark & \checkmark & 37.01 & ~~10.40 \\\hline
    \end{tabular}
    
    \caption{R@5 and MedR for the models with different inputs. The model uses action verb or pre-action image if it is checked. The first line denotes random search.}
    \label{tab:experimental-results}
\end{table}

\section{Application}\label{sec:application} 
\subsection{Multimodal commonsense reasoning}
Multimodal commonsense reasoning in recipe text is one of the recent trends~\citep{yagcioglu2018recipeqa,alikhani2019cite}. With our dataset, one can try reasoning about the food state changes from a raw ingredient to the final dish with the visual modality~\citep{bosselut2018simulating,nishimura2021state}. One can also use our dataset for analyzing the cooking action effects throughout a recipe.

\subsection{Procedural text generation}
Generating procedural text from vision is an important task~\citep{ushiku2017procedural,nishimura2019procedural}. To correctly reproduce procedures, the generated instructions should be consistent. The r-FG has the potential to make them more consistent as it represents the flow of the instructions. Since our recipes are associated with cooking videos, one can use our dataset for that purpose.

\section{Conclusion}\label{sec:conclusion}
We have presented a new multimodal dataset called Visual Recipe Flow. The dataset provides dense visual annotations for object states before and after a cooking action. The annotations allows us to learn each cooking action result. Experimental results demonstrated the effectiveness of our annotations for a multimodal information retrieval task. With our dataset, one can also try various applications, including multimodal commonsense reasoning and procedural text generation.

\section*{Acknowledgments}
We would like to thank anonymous reviewers for their insightful comments. This work was supported by JSPS KAKENHI Grant Number 20H04210, 21H04910, 22H00540, 22K17983 and JST PRESTO Grant Number JPMJPR20C2.

\bibliography{anthology,custom}
\appendix
\begin{table}[t]
    \begin{center}
        \begin{tabular}{l|l|r} \hline
            \multicolumn{1}{c|}{Tag} & \multicolumn{1}{c|}{Meaning} & \multicolumn{1}{c}{\# tags} \\\hline
            F & Food & 5,098 \\
            T & Tool & 758 \\
            D & Duration & 129 \\
            Q & Quantity & 1,778 \\
            \multirow{2}{*}{Ac} & \multirow{2}{*}{\makecell{Action by chef \\ (cooking action)}} & \multirow{2}{*}{2,532} \\
            & & \\
            Af & Action by food & 353 \\
            Sf & State of food & 971 \\
            St & State of tool & 67 \\\hline
            \multicolumn{1}{c|}{Total} & --- & 11,686 \\\hline
        \end{tabular}
    \end{center}
    \caption{r-NE tags and the number of annotated tags of each type.}
    \label{tab:r-NE-tags}
\end{table}
\begin{table}[t]
    \centering
    \scalebox{0.925}{
        \begin{tabular}{l|l|r} \hline
            \multicolumn{1}{c|}{Label} & \multicolumn{1}{c|}{Meaning} & \multicolumn{1}{c}{\# labels} \\\hline 
            Agent & Action agent & 330 \\
            Targ & Action target & 2,961 \\
            Dest & Action destination & 1,025 \\
            T-comp & Tool complement & 157 \\
            F-comp & Food complement & 20 \\
            F-eq & Food equality & 2,397 \\
            F-part-of & Food part-of & 330 \\
            F-set & Food set & 987 \\
            T-eq & Tool equality & 4 \\
            T-part-of & Tool part-of & 0 \\
            A-eq & Action equality & 1 \\
            V-tm & Head of clause for timing & 112 \\
            other-mod & Other relationships & 2,967 \\\hline
            \multicolumn{1}{c|}{Total} & --- & 11,291 \\\hline
        \end{tabular}
    }
    \caption{r-FG labels and the number of annotated labels of each type.}
    \label{tab:r-FG-labels}
\end{table}
\section{Detailed statistics for the textual annotation}\label{sec:r-NE-r-FG-stats}
This section provides the detailed statistics for the annotated r-NE tags and r-FG labels.

\subsection{r-NE tags}
\tabref{tab:r-NE-tags} shows the statistics for the annotated r-NE tags with the explanation of each tag. Among the tags, \textsf{Ac}, \textsf{F}, and \textsf{T} are specially important in our work. \textsf{Ac} denotes human cooking action, which is distinguished from action by food (\textsf{Af}). For example, in the instruction of ``leave the salad to cool,'' ``leave'' is tagged with \textsf{Ac}, while ``cool'' is tagged with \textsf{Af}. \textsf{F} denotes foods including raw ingredients, intermediate products after cooking action, and the final dish. \textsf{T} denotes tools used for cooking. In our work, objects refer to the foods or tools. Our image annotation targeted the states of these objects.

\subsection{r-FG labels}
\tabref{tab:r-FG-labels} shows the statistics for the annotated r-FG labels with the explanation of each tag. The cooking action (\textsf{Ac}) requires the objects (\textsf{F} or \textsf{T}). \textsf{Targ} describes this relationship taking the action and object as the starting and ending vertices, respectively. During the image annotation, we identified the required objects by using the flows labeled with \textsf{Targ}.

\begin{figure*}[t]
    \centering
    \includegraphics[scale=0.4585]{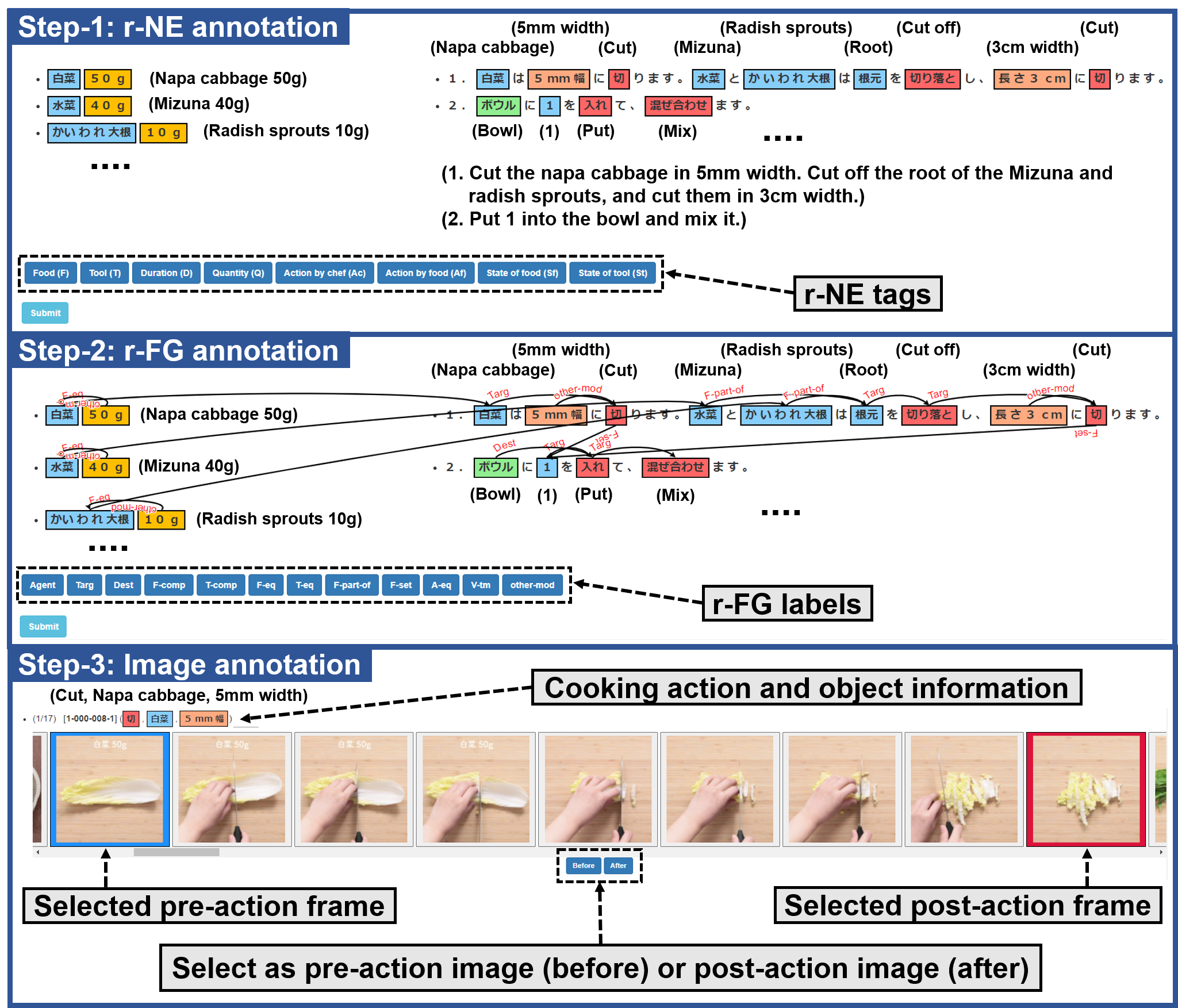} 
    \caption{Our web annotation interface. The annotator can complete annotations only by mouse operations. The web page is written in Japanese.}
    \label{fig:web-interface}
\end{figure*}

\section{Web interface}\label{sec:web-interface}
Our developed web interface is illustrated in \figref{fig:web-interface}. In the first step (r-NE annotation), the annotator can annotate words in the ingredient and instruction lists with an r-NE tag by clicking the words and tag. In the second step (r-FG annotation), the annotator can annotate the r-NEs with a recipe flow by clicking starting and ending vertices and a label for them. In the final step (image annotation), the annotator can annotate the pre-action and post-action object states with images by clicking a frame and the button for the state. All objects for annotation are automatically prepared by tracing the recipe flows.

\section{A joint embedding model}\label{sec:appendix-model}
In this section, we provide the detailed calculation of our model and experimental settings.

\subsection{Model description}\label{subsec:joint-embedding-model}
We first calculate a vector for an estimated post-action object state baesd on an action verb $a$, an object information $o$, and a pre-action image $i_{pre}$. The object is obtained by tracing a recipe flow labeled with \textsf{Targ}. $a$ and $o$ are converted to $d_t$-dimensional vectors $h_a$ and $h_o$, respectively, by first embedding words into $d_v$-dimensional representations via a lookup table and then encoding them into $d_t$-dimensional vectors by using a bidirectional LSTM (BiLSTM)~\citep{graves2005framewise}. For $i_{pre}$, we extract its feature $h^{pre}_{i} \in \mathbb{R}^{d_i}$ by using a pre-trained convolutional neural network (CNN) and transform it into $\hat{h}^{pre}_{i} \in \mathbb{R}^{d_t}$ as follows:
\begin{equation}
    \hat{h}^{pre}_{i} = W^{T}_{1}h^{pre}_{i} + b^{T}_{1}, \label{equ:image-transform}
\end{equation}
where $W^{T}_{1} \in \mathbb{R}^{d_t \times d_i}$ and $b^{T}_{1} \in \mathbb{R}^{d_t}$ are learnable parameters. Given these fixed-size vectors, we then compute the vector for the estimated post-action object state $\hat{h}_{o}$ as:
\begin{equation}
    \hat{h}_{o} = W^{T}_{3}(\textrm{ReLU}(W^{T}_{2}[h_a;h_o;h^{pre}_{i}] + b^{T}_{2})) + b^{T}_{3}, \label{equ-text-encoding}
\end{equation}
where $;$ denotes concatenation, and $W^{T}_{2} \in \mathbb{R}^{3d_t \times 3d_t}$, $W^{T}_{3} \in \mathbb{R}^{d_t \times 3d_t}$, $b^{T}_{2} \in \mathbb{R}^{3d_t}$, and $b^{T}_{3} \in \mathbb{R}^{d_t}$ are learnable parameters. $\hat{h}_{o}$ is then mapped to the joint embedding space as:
\begin{eqnarray}
    h_t & = & (W^{T}_{4}\hat{h}_{o} + b^{T}_{4}) \circ \nonumber \\
    & & \sigma(W^{T}_{5}(W^{T}_{4}\hat{h}_{o} + b^{T}_{4}) + b^{T}_{5}), \label{equ-text-mapping}\\
    \tilde{h}_t & = & \frac{h_t}{||h_t||_2}, \label{equ-text-l2}
\end{eqnarray}
where $W^{T}_{4} \in \mathbb{R}^{d_e \times d_t}$, $W^{T}_{5} \in \mathbb{R}^{d_e \times d_e}$, $b^{T}_{4},b^{T}_{5} \in \mathbb{R}^{d_e}$ are learnable parameters. 

The post-action image $i_{post}$ is fed to the pre-trained CNN to extract its feature $h^{post}_{i} \in \mathbb{R}^{d_i}$. Based on this feature, we compute $\hat{h}_{i}$ as:
\begin{equation}
    \hat{h}_{i} = W^{I}_{2}(\textrm{ReLU}(W^{I}_{1}h^{post}_{i} + b^{I}_{1})) + b^{I}_{2}, \label{equ-image-encoding}
\end{equation}
where $W^{I}_{1},W^{I}_{2} \in \mathbb{R}^{d_i \times d_i}$, and $b^{I}_{1},b^{I}_{2} \in \mathbb{R}^{d_i}$ are learnable parameters. Following \citet{miech2018learning}, the feature vector $\hat{h}_{i}$ is then mapped to the joint embedding space as follows:
\begin{eqnarray}
    h_v & = & (W^{I}_{3}\hat{h}_{i} + b^{I}_{3}) \circ \nonumber \\
    & & \sigma(W^{I}_{4}(W^{I}_{3}\hat{h}_{i} + b^{I}_{3}) + b^{I}_{4}), \label{equ-image-mapping}\\
    \tilde{h}_v & = & \frac{h_v}{||h_v||_2}, \label{equ-image-l2}
\end{eqnarray}
where $\sigma$ is the sigmoid function, $\circ$ denotes the element-wise multiplication, $W^{I}_{3} \in \mathbb{R}^{d_e \times d_i}$, $W^{I}_{4} \in \mathbb{R}^{d_e \times d_e}$, and $b^{I}_{3},b^{I}_{4} \in \mathbb{R}^{d_e}$ are learnable parameters. 

\paragraph{Loss function.}
After mapping the inputs to the joint embedding space, we calculate the distance between these vectors as:
\begin{equation}
    D(\tilde{h}_{t}, \tilde{h}_{v}) = ||\tilde{h}_{t} - \tilde{h}_{v}||_2.
\end{equation}
Given $n$ examples of ($(\tilde{h}_{t,1}$,$\tilde{h}_{v,1}), \cdots, (\tilde{h}_{t,n}, \tilde{h}_{v,n})$), we minimize the following triplet loss~\citep{balntas2016learning}:
\begin{eqnarray}
    \mathcal{L} = \sum^{n}_{i=1} & \{ \max(D_{i,i} - D_{i,j} + \delta, 0) \nonumber \\
    & + \max(D_{i,i} - D_{k,i} + \delta, 0) \},
    \label{equ-triplet-loss}
\end{eqnarray}
where $D_{i,j} = D(\tilde{h}_{t,i}, \tilde{h}_{v,j})$, and $\delta$ denotes a margin. In \equref{equ-triplet-loss}, $D_{i,i}$ is the distance for a positive pair, and $D_{i,j}$ and $D_{k,i}$ are the distances for pairs with negative text and image feature vectors, respectively. For negative sampling, we simply sample negative examples from a mini-batch.

\subsection{Settings}
\paragraph{Model parameters.} We used a 1-layer $256$-dimensional BiLSTM to encode words. We set the dimensions as $(d_v, d_t, d_i, d_e) = (496, 512, 2048, 128)$. We used ResNet-152~\citep{he2016deep}, which was pre-trained on ImageNet~\citep{russakovsky2015imagenet}, to extract a feature vector of $2048$ dimensions from an image.

\paragraph{Optimization.} We used AdamW~\citep{loshchilov2019decoupled} with an initial learning rate of $1.0 \times 10^{-5}$ to tune the parameters. During training, we froze only the parameters of the CNN. Each model was trained for $350$ epochs, and we created a mini-batch with $4$ recipes at each step. We set $\delta$ in \equref{equ-triplet-loss} to $0.1$.  We evaluated the model performance through 10-fold cross-validation by splitting the dataset into 90\% for training and 10\% for testing.



\end{document}